\documentclass[10pt,twocolumn,letterpaper]{article}

\usepackage{cvpr}
\usepackage{times}
\usepackage{epsfig}
\usepackage{graphicx}
\usepackage{amsmath}
\usepackage{amssymb}

\usepackage{subcaption}
\usepackage{multirow}
\usepackage{float}
\usepackage[inline]{enumitem}
\usepackage{color}

\usepackage[pagebackref=true,breaklinks=true,letterpaper=true,colorlinks,bookmarks=false]{hyperref}

\cvprfinalcopy 

\def\cvprPaperID{61} 

\ifcvprfinal\fi
\begin{document}

\title{Suppressing Model Overfitting for Image Super-Resolution Networks}

\author{Ruicheng Feng$^1$, Jinjin Gu$^2$, Yu Qiao$^{1, 3}$, Chao Dong$^1$\\
$^1$ShenZhen Key Lab of Computer Vision and Pattern Recognition, SIAT-SenseTime Joint Lab,\\
Shenzhen Institutes of Advanced Technology, Chinese Academy of Sciences\\
$^2$The School of Science and Engineering, The Chinese University of Hong Kong, Shenzhen\\
$^3$The Chinese University of Hong Kong\\
{\tt\small \{rc.feng, yu.qiao, chao.dong\}@siat.ac.cn, jinjingu@link.cuhk.edu.cn}
}

\maketitle

\begin{abstract}
Large deep networks have demonstrated competitive performance in single image super-resolution (SISR), with a huge volume of data involved.
However, in real-world scenarios, due to the limited accessible training pairs, large models exhibit undesirable behaviors such as overfitting and memorization.
To suppress model overfitting and further enjoy the merits of large model capacity, we thoroughly investigate generic approaches for supplying additional training data pairs. 
In particular, we introduce a simple learning principle \textit{MixUp} \cite{zhang2017mixup} to train networks on interpolations of sample pairs, which encourages networks to support linear behavior in-between training samples.
In addition, we propose a data synthesis method with learned degradation, enabling models to use extra high-quality images with higher content diversity. This strategy proves to be successful in reducing biases of data.
By combining these components -- MixUp and synthetic training data, large models can be trained without overfitting under very limited data samples and achieve satisfactory generalization performance. Our method won the second place in NTIRE2019 Real SR Challenge.
\end{abstract}

\section{Introduction}
\begin{figure}[t]
\begin{center}
   \includegraphics[width=1.0\linewidth]{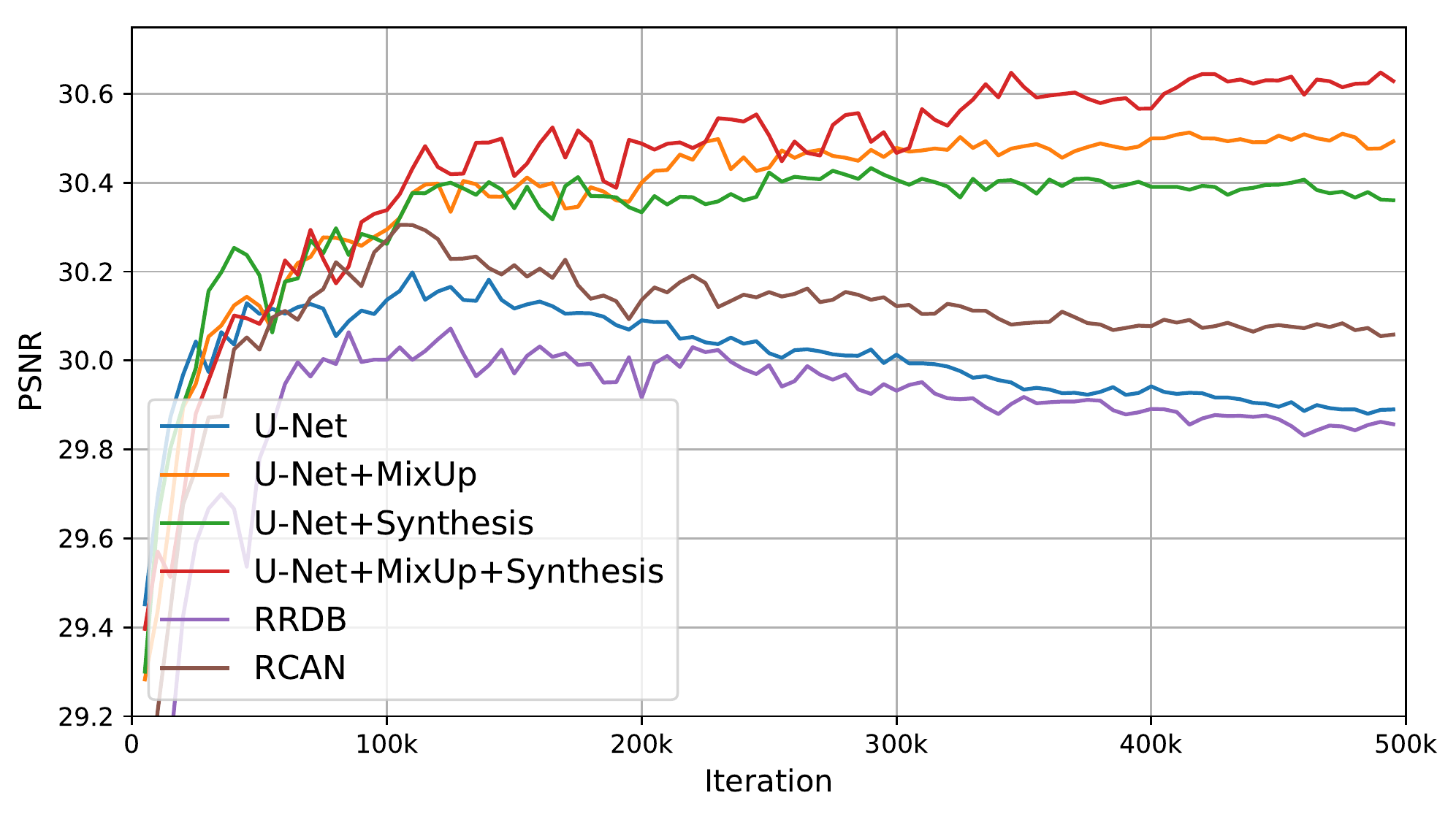}
\end{center}
\vskip -0.6cm
\caption{Convergence curves of RRDB\cite{wang2018esrgan}, RCAN\cite{zhang2018image}, the proposed U-Net and its variants with different data augmentation techniques. The original large models suffer from different degrees of overfitting, while same models trained with either MixUp, data synthesis, or both can achieve satisfactory performance without overfitting.}
\vskip -0.6cm
\label{fig:comparison}
\end{figure}

Since the seminal work of employing convolution neural networks (CNNs) for single image super-resolution (SISR) \cite{dong2014learning,dong2016image}, a constantly growing flow of deep learning based methods with different network architectures \cite{dong2016accelerating,kim2016accurate, lai2017deep,kim2016deeply,tai2017memnet,haris2018deep,zhang2018residual,zhang2018image,ahn2018fast} and training strategies \cite{wang2018esrgan, shocher2018zero, bulat2018learn,gu2019blind,qian2019trinity} have been proposed to achieve substantial progress in state-of-the-art performance.
These methods are usually trained and tested using thousands of high-quality images.
Therefore, overfitting is rarely observed when training models with such abundant image pairs.
These image pairs are usually generated by pre-defined downsampling methods, such as bicubic.
Beyond those pre-defined degraders, in the recent work \cite{chen2019camera,cai2019toward,Zhang_2019_CVPR} real captured low-high resolution image pairs are used to train SR models under realistic application settings.
However, the amount of such data is often limited (e.g., only $60$ image pairs in NTIRE19 Real SR Challenge \cite{ntire2019}) because of the high cost of collection and preprocessing of data.
This leads to severe overfitting problem for recent deep SR networks.
Specifically, the network tends to memorize the training images and generalizes poorly to the test set. 
For instance, as shown in \figurename~\ref{fig:comparison}, large models trained on a small dataset quickly deteriorate their generalization performance (see the lower curves).
The overfitting problem has largely limited the usage of the advanced SR methods in real-world applications.

As an important issue, overfitting has attracted increasingly research interests in high-level vision tasks, such as image classification \cite{devries2017improved,geirhos2018imagenet,inoue2018data,cubuk2018autoaugment,wang2018data}, visual tracking \cite{danelljan2016adaptive,gao2014transfer}, etc.
However, overfitting in low-level tasks has received relatively less attention.
Due to the different characteristics of low-/high-level tasks, most existing methods that are suitable for high-level tasks cannot be directly applied to low-level tasks.
For example, some network regularization methods, such as weight decay and dropout, do not work effectively for low-level networks.
In addition, some popular data augmentation techniques such as label smoothing are also infeasible for low-level tasks as they only work with one-hot labels.
In low-level vision community, only limited augmentation methods (e.g., random crop, rotation and flipping) are investigated, which is far from sufficiency for real-world applications.

In this paper, we study the overfitting problem for SR.
First, we adopt a simple yet effective data augmentation method called \textit{MixUp} \cite{zhang2017mixup} in SR.
MixUp uses convex combinations of samples rather than samples themselves to train the SR model.
It normalizes neural networks to support simple linear behavior in-between training samples, and leads to better generalization performance (see orange curve in \figurename~\ref{fig:comparison}).
Second, we propose a data synthesis approach with a learned degradation mapping.
Concretely, we use deep networks to learn the degradation mapping first, and synthesize new training samples using extra high-quality images.
This synthesis strategy reduces the bias of the data by introducing content diversity into the training set (see green curve in \figurename~\ref{fig:comparison}).
The SR models trained with the synthetic data are expected to provide better generalization performance on image contents that do not exist in the original small dataset.
By combining the above components -- MixUp and synthetic training data, we are able to suppress model overfitting in SR under very limited training samples.
Extensive experiments show that either MixUp, data synthesis, or both can suppress model overfitting and encourage better generalization (see upper curves in \figurename~\ref{fig:comparison}).

We summarize our contributions as follows:
(1) We introduce the MixUp technique into SR for data augmentation. Experiments demonstrate that MixUp could significantly reduce the overfitting problem.
(2) We propose a new data synthesis method to suppress model overfitting in SR. It uses the learned degradation mapping to synthesize more training pairs with additional high-quality images.
(3) With the proposed data augmentation and data synthesis methods, we construct a network of a general U-Net shape \cite{ronneberger2015u} which encourages better generalization ability and achieves satisfactory performance without overfitting. Our method won the second place in NTIRE 2019 Real SR Challenge.

\section{Related Work}
\textbf{Image super-resolution}
Recently, learning-based methods have achieved dramatic advantages against the model based methods.
With the seminal exploration of employing deep learning in SR task \cite{dong2014learning,dong2016image}, the variational approaches with deep neural networks have been dominated single image SR.
Dong \textit{et al.} \cite{dong2016accelerating} propose to use a deeper network with low-resolution image as input to learn the SR mapping.
Kim \textit{et al.} \cite{kim2016accurate} propose VDSR -- a very deep network with residual learning and show the performance improvement by using deep networks.
Ledig \textit{et al.} \cite{ledig2017photo} introduce residual blocks into SR network and propose SRResNet, which makes it possible to train deeper networks.
Lim \textit{et al.} \cite{lim2017enhanced} further expand the network size and improve the residual block by removing the Batch Normalization Layers.
Zhang \textit{et al.} \cite{zhang2018image} propose a deep network with dense connection and Wang \textit{et al.} \cite{wang2018esrgan} propose to use residual in residual dense block to improve the training stability and network size.
Zhang \textit{et al.} \cite{zhang2018residual} propose residual channel attention blocks and indicate that deeper networks may be easier to achieve better performance than wider networks.
As can be seen, most recently successful SR methods employ very deep networks with a large number of parameters, which leads to a high risk of overfitting.

\textbf{Data augmentation.}
The method of choice to train on similar but different examples to the training data is known as data augmentation \cite{simard1998transformation}.
The most common methods of data augmentation include some basic image processing operations, e.g., random scale, random crop, horizontal/vertical flip and image affine transformation.
In addition to the basic image processing operations, Zhong \textit{et al.} \cite{zhong2017random} propose to augment data by randomly erasing part of the image.
Inoue \cite{inoue2018data} propose to synthesize a new sample from one image by overlaying another image randomly chosen from the training data.
Zhang \textit{et al.} \cite{zhang2017mixup} propose to synthesize new samples using the linear combination of training samples.
DeVries \textit{et al.} \cite{devries2017improved} improves regularization of networks by masking out square region of training images.
Geirhos \textit{et al.} \cite{geirhos2018imagenet} reduces bias toward textures by introducing stylized image data for training.
Cubuk \textit{et al.} \cite{cubuk2018autoaugment} presents AutoAugment to learn the best augmentation policies from data.
Besides, Generative adversarial networks (GANs) have also been used for the purpose of generating additional data \cite{perez2017effectiveness, mun2017generative, zhu2017data, antoniou2017data,sixt2018rendergan,ratner2017learning}.
Most of the existing data augmentation methods are proposed and studied for high-level tasks, and there exists few work to study the effects of different data augmentation methods on the low-level task such as SR.

\textbf{NTIRE 2019 Real Super-Resolution Challenge.}
This work is initially developed to participate in the NTIRE2019 Real Super-Resolution Challenge \cite{ntire2019}.
The challenge aims to offer an opportunity for academic and industrial attendees to focus on Super-Resolution applications in real-world scenario.
In the challenge, a novel dataset of LR real images with HR real references, where the sizes of LR images are same as its HR counterparts, is provided to challenge participants.
These images are collected in natural environments, including indoor and outdoor environments.
Different from most SISR tasks \cite{dong2016image, lim2017enhanced} using pre-defined degraders, images from this dataset are captured by DSLR cameras, and therefore facilitate researches for real-world applications.

However, due to the small volume of data pairs, models suffer from severe overfitting problem.
Hence, mechanisms for training large models without overfitting are required to deal with this challenge.
We submitted our models and prove that our method are able to suppress model overfitting in SR.
Our methods successfully reconstruct HR images from severely degraded real LR images without unpleasant artifacts related to overfitting.
Our approach won the second place in the challenge.

\section{Methodology}
In this section we show the overfitting problem in SR and present our proposed methods.
The rest of this section is organized as follows: Sec. \ref{overfit} describes how SR networks overfit on training dataset from NTIRE 2019 Real SR Challenge.
Then, we formulate the overfitting issue and data augmentation.
Later, Sec. \ref{mixup} and \ref{syn} introduce the data augmentation method with MixUp and the data synthesis method with learned degradation, respectively.
Finally, in Sec. \ref{arch} we illustrate the network architecture.

\subsection{Overfitting in Super Resolution}
\label{overfit}

\begin{figure}[t]
\centering
    \begin{subfigure}[t]{1.0\linewidth}
        \centering
        \includegraphics[width=0.95\linewidth]{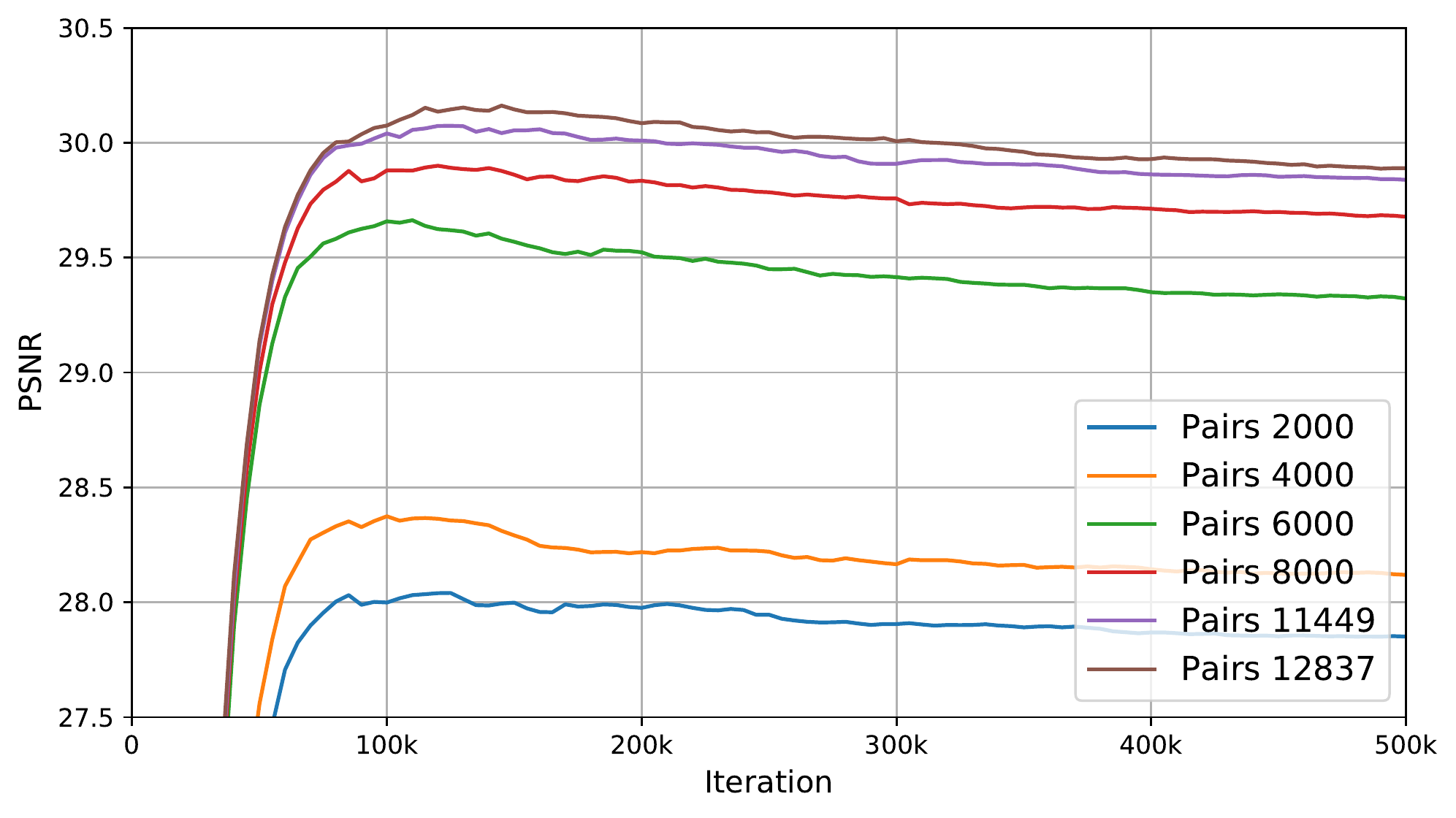}
        \caption{Convergence curves of models trained with different amounts of data.}
        \label{fig:ccdata}
    \end{subfigure}\\
    \begin{subfigure}[t]{1.0\linewidth}
        \centering
        \includegraphics[width=0.95\linewidth]{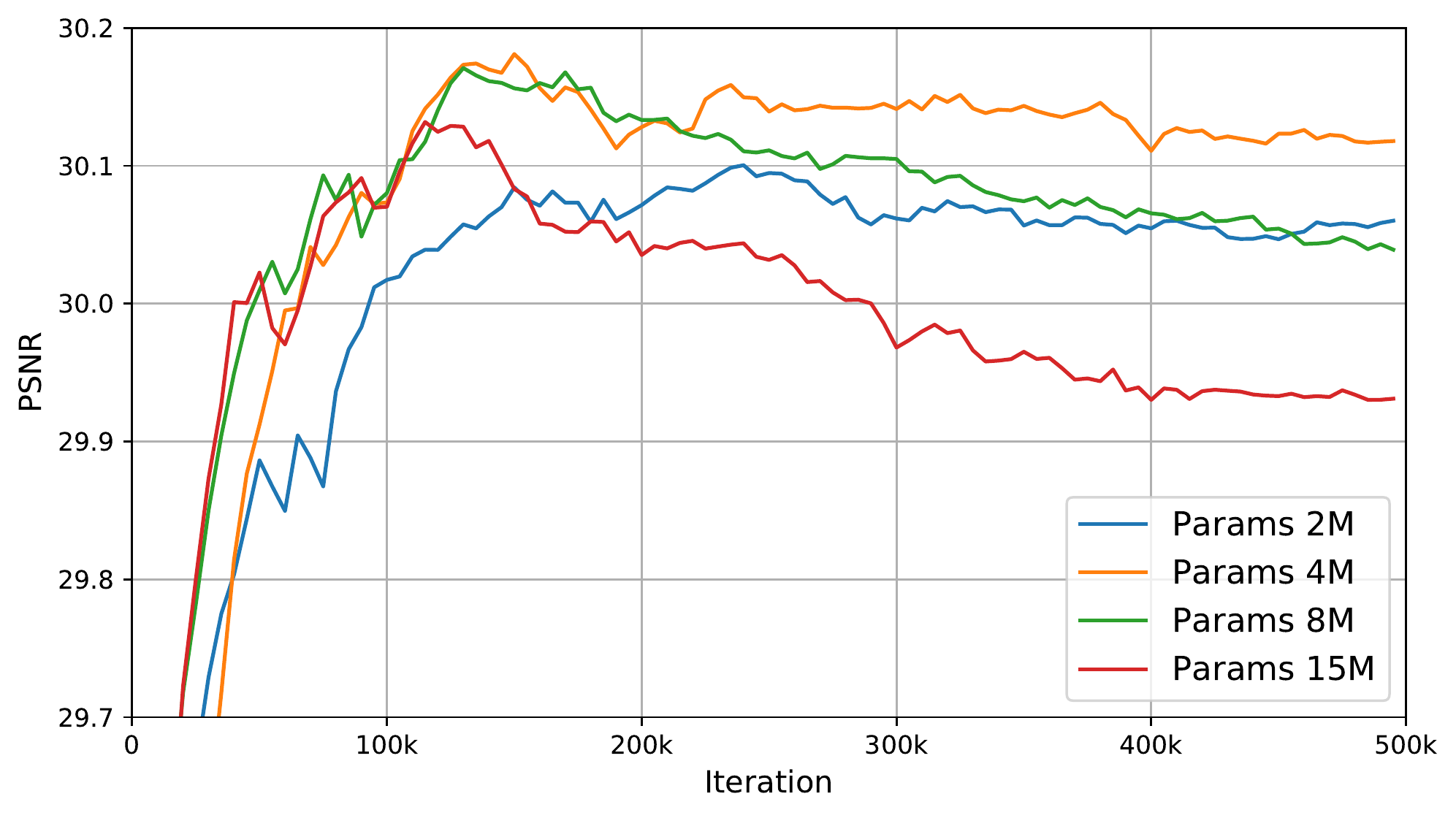}
        \caption{Convergence curves of models with different complexities.}
        \label{fig:ccmodel}
    \end{subfigure}
\caption{Illustration on impact of amounts of data and model complexities on validation performance.}
\vskip -0.6cm
\label{fig:overfitting}
\end{figure}

In this challenge, a new dataset of real LR and HR paired images (RealSR), with the spatial resolution no smaller than $1000\times1000$, is publicly available.
This dataset contains only $60$ images for training (See Sec. \ref{dataset} for details).
Due to the limited diversity and amount of training data, large models exhibit undesirable overfitting behaviors even when using straightforward data augmentation techniques (e.g. random crop, rotation, flipping).
For instance, a well-trained model poorly generalize to the test set and tends to generate unpleasant artifacts on test images.

To start off with right intuitions, \figurename~\ref{fig:overfitting} illustrates the impact of data volume and model complexity evaluated on the validation set.
The validation set consists of $20$ images covering contents that do not exist in the training set.
In the first setting, we construct a sufficiently large network (with $26$M parameters) and train the network with different sizes of data, starting with the first $2,000$ sub-images (from about $10$ images) and increasing gradually to all $12,837$ sub-images (cover $60$ images). 
In \figurename~\ref{fig:ccdata}, we can observe that while all models quickly overfit to training set, increasing amounts of training data will lead to better performance in the training phase.
In another setting, we use the whole training set to train models with different sizes, ranging from $2$M to $15$M.
\figurename~\ref{fig:ccmodel} shows that larger models do not necessarily achieve higher PSNR values at the early stage and suffer from severe overfitting if training continues.
In contrast, the overfitting problem on small models becomes less severe.
This example conveys the central message: overfitting in SR is partially due to the mismatch between data volume and model complexity.
To enjoy the merits of large model, we present two methods to remedy such a discrepancy by supplying additional training pairs.

\subsection{Problem Formulation}
\begin{figure*}[t]
\centering
    \begin{subfigure}[t]{0.3\linewidth}
        \centering
        \includegraphics[width=0.9\linewidth]{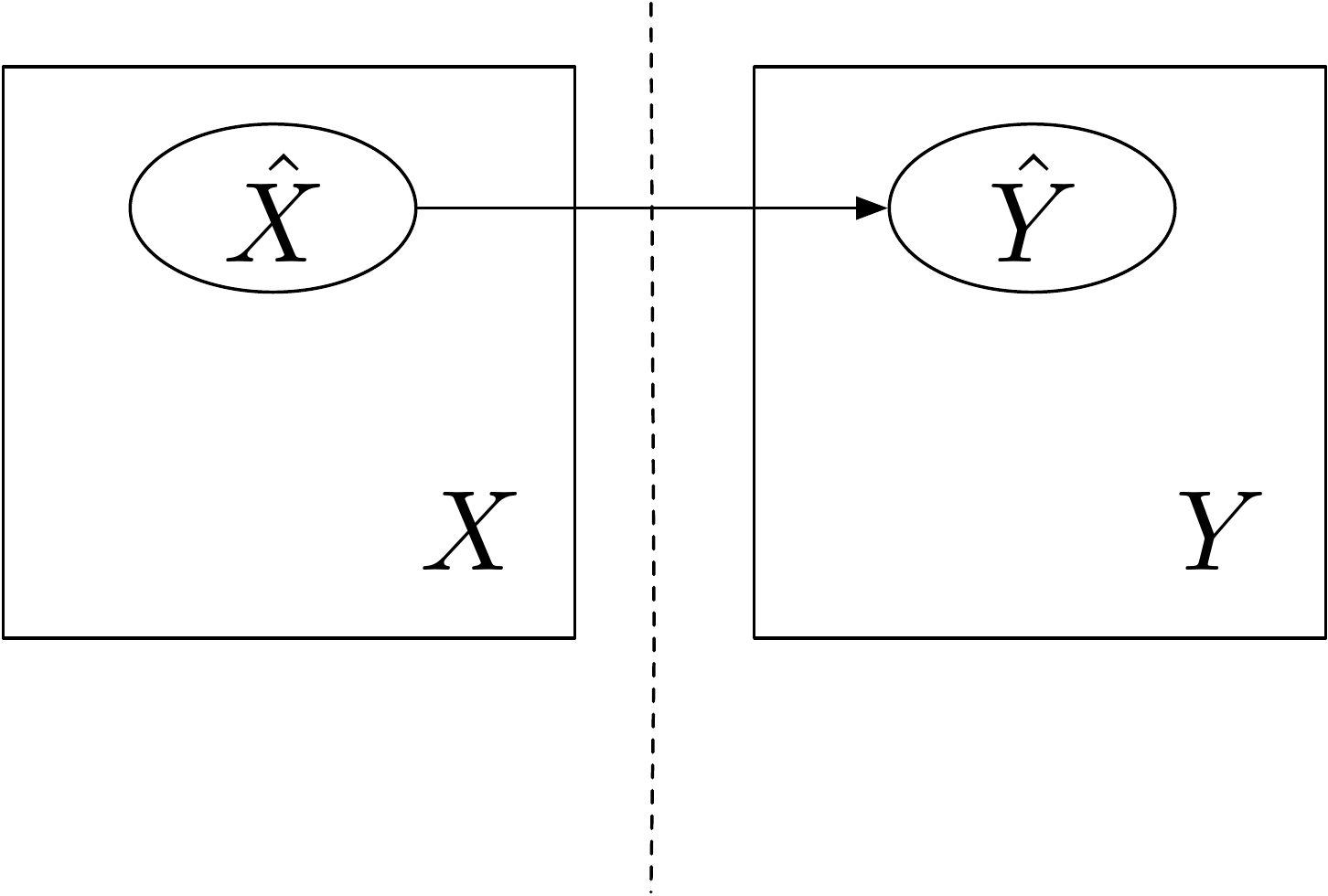}
        \caption{Original SISR.}
        \label{fig:can1}
    \end{subfigure}
    \begin{subfigure}[t]{0.3\linewidth}
        \centering
        \includegraphics[width=0.9\linewidth]{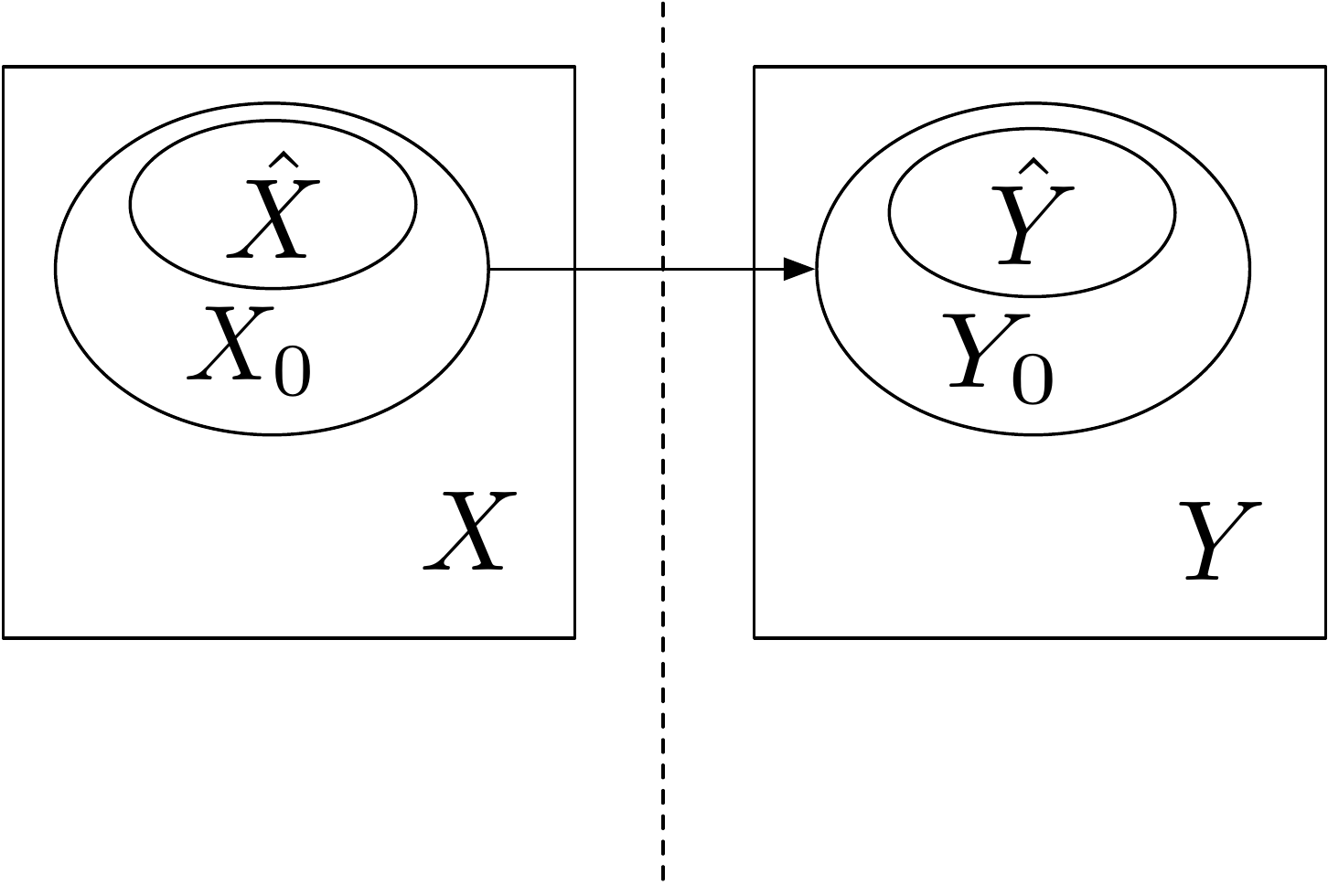}
        \caption{MixUp augmentation.}
        \label{fig:can2}
    \end{subfigure}
    \begin{subfigure}[t]{0.3\linewidth}
        \centering
        \includegraphics[width=0.9\linewidth]{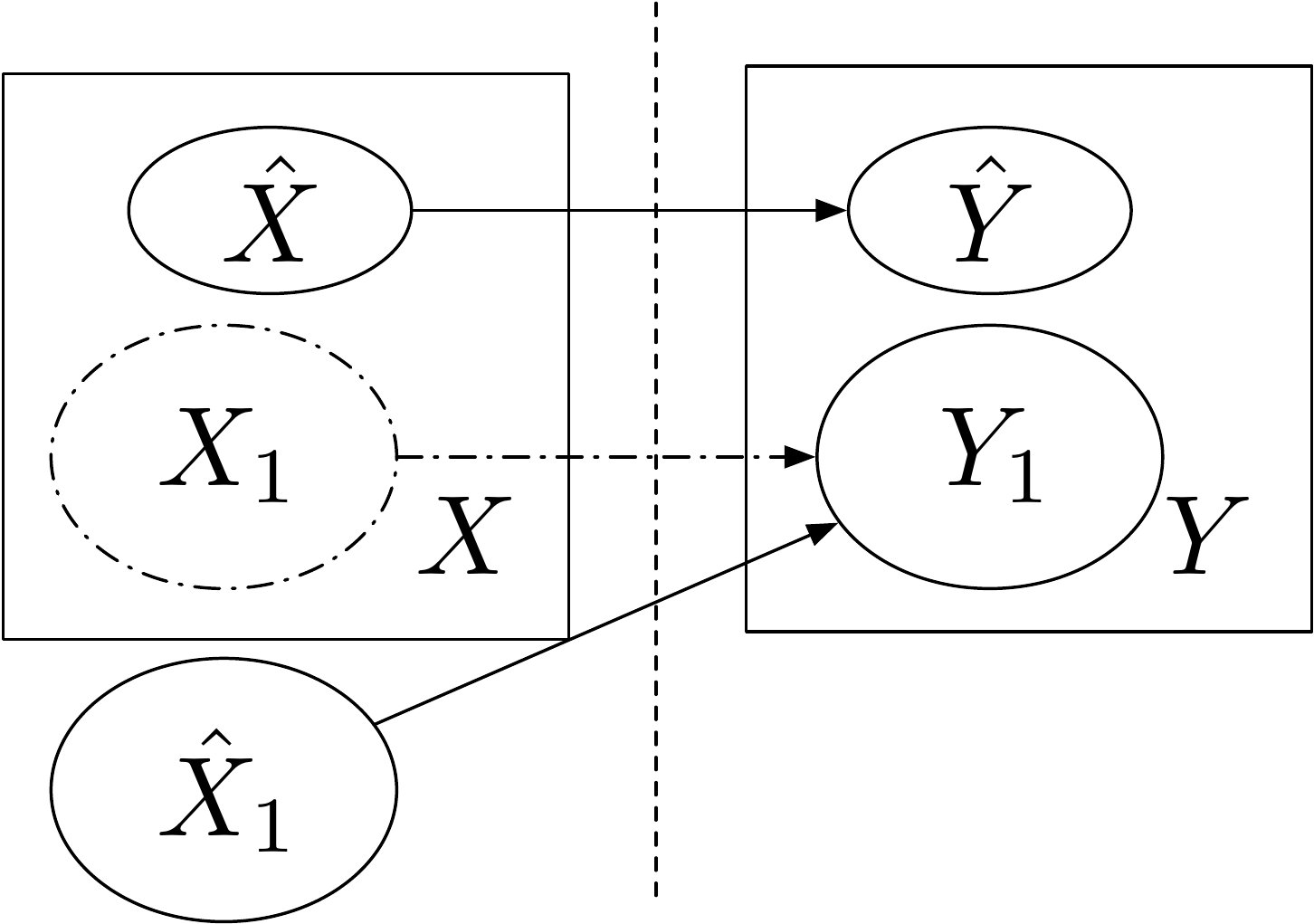}
        \caption{Data synthesis.}
        \label{fig:can3}
    \end{subfigure}
    \vskip -0.2cm
\caption{Illustration on how data augmentation and data synthesis work. (a) The observation set $(\hat X,\hat Y)$ is a subset of the true data set $(X,Y)$. (b) MixUp technique supplies additional training pairs and the augmentation set $(X_0,Y_0)$ covers the observation set. (c) Data synthesis method estimates inaccessible LR images $X_1$ from extra high-quality HR images $Y_1$. The estimation $\hat X_1$, accompanied with $Y_1$, constitutes a synthetic dataset and help to reduce the risk of overfitting.}
\vskip -0.4cm
\label{fig:overfitting}
\end{figure*}

To facilitate the discussion, we first formulate the overfitting problem and data augmentation.
Let $X$, $Y$ be the LR images and their HR counterparts on the true data space, where true data refer to image pairs with the desired degradation function, which can be either pre-defined kernels or unknown real degradations. For each $y\in Y$, we have $x=g(y)$, where $g$ is the \textit{degradation} function mapping $Y$ onto $X$.
In SISR task, given an observation set $(\Hat{X}, \Hat{Y})\subset(X,Y)$ as the training set, our goal is to find an inverse mapping function $f_{\theta}$ by optimizing a well-defined loss function $\mathcal{L}$
\begin{equation}
\label{mle}
    \Hat{\theta} = \mathop{\arg\min}_{\theta} \sum_{(x,y)\in(\Hat{X}, \Hat{Y})}\mathcal{L}(f_{\theta}(x),y).
\end{equation}
The major risk of this framework is that $f_{\theta}$ may be \textit{biased}, leading to poor generalization ability on unobserved data points.
This problem is severe especially when observations are insufficient to cover the true data manifold.

The most widely-used technique to reduce such a risk is data augmentation.
Specifically, in the perspective of data augmentation, an addition set $(X',Y')$, which is beyond the training set $(\Hat{X}, \Hat{Y})$ but believed inside the true data manifold $(X,Y)$, are introduced for training.
In SISR, $(X',Y')$ can be obtained by rotating each data pair in $(\Hat{X}, \Hat{Y})$.
We hypothesize that for each $(x,y)\in (X',Y')$, we have $g(y)=x$, indicating that data pairs in observation set and those in augmentation set follow the same degradation mapping.

\subsection{Data Augmentation with MixUp}
\label{mixup}
We consider a simple yet effective data augmentation method, \textit{MixUp} \cite{zhang2017mixup}.
In MixUp, each time we randomly sample two samples $(x_i , y_i )$ and $(x_j , y_j)$ in the set $(\Hat{X}, \Hat{Y})$.
Then we form a new sample by a linear interpolation of these two samples:
\begin{align}            
	x'&=\lambda x_i + (1 - \lambda) x_j\\
	y'&=\lambda y_i + (1 - \lambda) y_j,
\end{align}
where $\lambda\in[0,1]$ is a random number drawn from a beta distribution $\mathbf{Beta}(\alpha, \alpha)$.

In super resolution, we can assume that the degradation function $g$ is a linear mapping, which can be formulated as $x=g(y)=Dy+n$, where $D$ is the downsampling matrix and $n$ is the noise.
If $D$ and $n$ are determinded, we have
\begin{align}
    x'&=\lambda x_i + (1 - \lambda) x_j\\
      &=\lambda (Dy_i + n_i)) + (1 - \lambda) (Dy_j + n_j)\\
      &=D(\lambda y_i + (1 - \lambda) y_j) + (\lambda n_i + (1-\lambda) n_j)\\
      &=Dy'+n',
\end{align}
where $n'=\lambda n_i+(1-\lambda) n_j$.
$n'$ is the noise and drawn from the same distribution of $n$. This property also holds when $n$ is signal-dependent.
This indicates that although the MixUp-augmented data pairs have unnatural visual effects, they follow the same degradation model with the true data and can be used to learn the inverse mapping $f$.

Moreover, MixUp provides a linear neighbourhood of real data, making the learned inverse mapping more robust.
With MixUp, we can easily obtain multiple times of data pairs to train the network. As illustrated in \figurename~\ref{fig:can2}, the observation set $(\Hat{X}, \Hat{Y})$ is a subset of MixUp-augmented dataset $(X_0, Y_0)$ and the latter on has greater cardinality.

Experiments in Sec. \ref{exp:mixup} show that this simple augmentation method can simultaneously suppress overfitting and improve performance.

\subsection{Data Synthesis with Learned Degradation}
\label{syn}
\begin{figure}[t]
\begin{center}
   \includegraphics[width=0.9\linewidth]{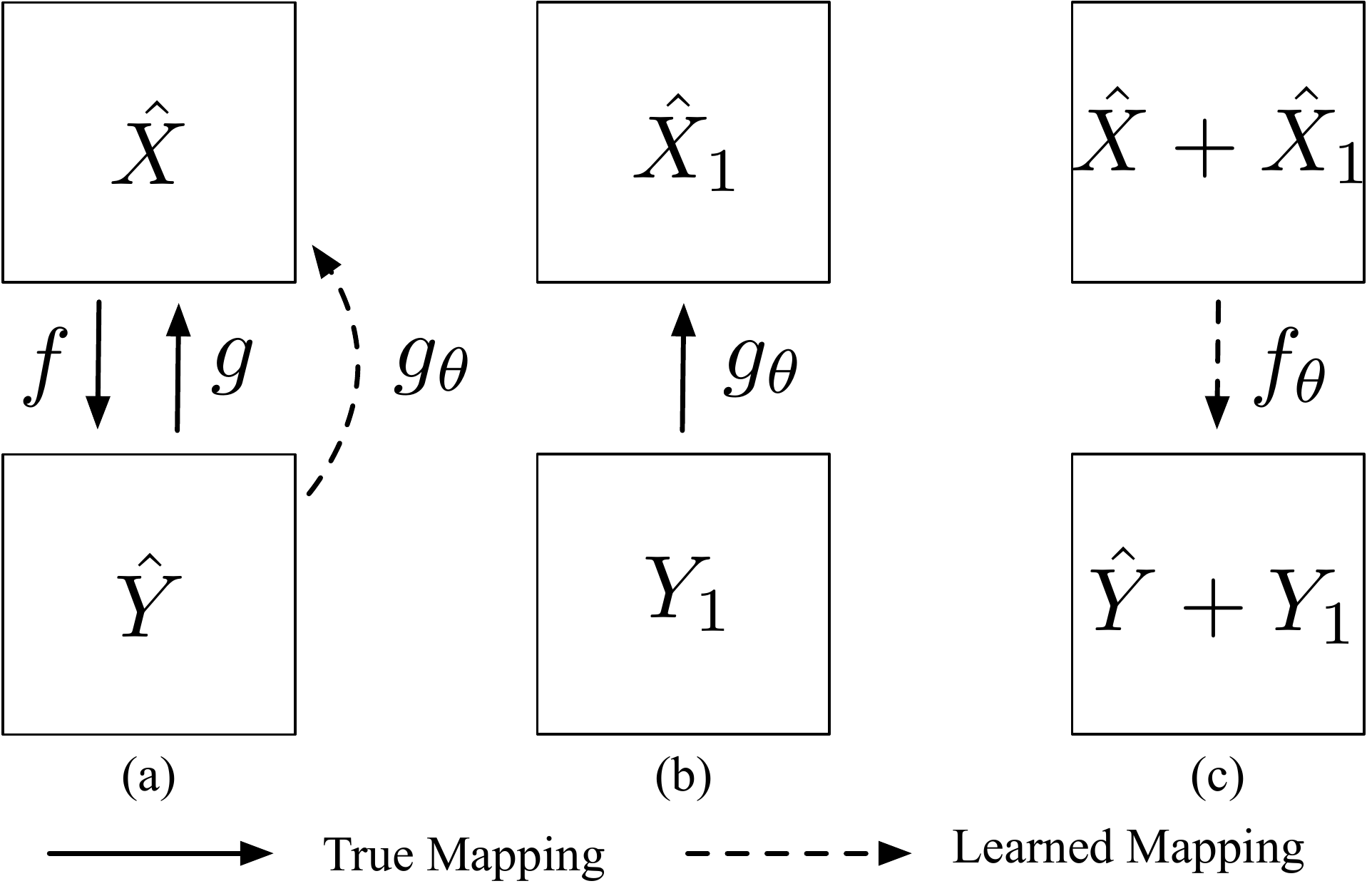}
\end{center}
\vskip -0.5cm
   \caption{Overview of pipeline for data synthesis. The approach aims to learn two mapping functions $g_{\theta}:\Hat{Y}\to \Hat{X}$ and $f_{\theta}:\Hat{X} +\Hat{X}_1\to \Hat{Y}+Y_1$. (a) Learn $g_{\theta}$ that $g_{\theta}(y) \approx g(y)$ and (b) synthesize LR images from $Y_1$. (c) The vanilla training process on both observed and synthetic data.}
\label{fig:addis}
\vskip -0.5cm
\end{figure}

Beyond MixUp, we also investigate another strategy to provide more training examples -- data synthesis via learning degradation process.
As depicted in \figurename~\ref{fig:addis}, given an observation set $(\Hat{X}, \Hat{Y})$ comprising images with finite content diversity, there might be a risk of biased sampling from the true data distribution.
Formally, let $\Hat{P}$ and $P$ be the observed and true data distribution, respectively.
For some training pairs $(x,y)\in(X,Y)$ with biased sampling, $\Hat{P}(x,y)$ could diverge far from $P(x,y)$.
In the extreme, suppose that there is an imbalanced training set with purely text images, then it is unlikely for models trained with such a dataset to generalize well on other contents (e.g., human face, natural scenery, animal, etc.).
In practice, a small set $(\Hat{X}, \Hat{Y})$ is usually both imbalanced and noisy, which increase the risk of overfitting.

To bridge the gap between $\Hat{P}$ and $P$, we propose a data synthesis technique to provide training pairs with higher diversity.
As illustrated in \figurename~\ref{fig:addis}, given a high-quality diverse HR dataset (e.g. DIV2K \cite{Agustsson_2017_CVPR_Workshops}, Flickr2K \cite{timofte2017ntire}, etc.) as $\Hat{Y}$, the corresponding LR image set $X_1$ is not accessible since the true degradation $g:Y\to X$ is unknown.
Due to nuisance factors, including blur (e.g. motion or defocus), compression artifacts, color and sensor noise, etc., it is usually impractical to effectively model the true image degradation in real-world scenarios.
Rather than managing to model a complicated image degradation process, we propose to use a neural network model denoted as $g_{\theta}$ to learn the degradation $g$ on finite observation set $(\Hat{Y}, \Hat{X})$.

With well-optimized $g_{\theta}$, we can obtain estimated LR images $\Hat{X}_1$, where for each $\Hat{x}\in\Hat{X}_1$ we have $\Hat{x} = g_{\theta}(y)$ for $y\in Y_1$.
As $g_{\theta}$ is an approximation of $g$, we expect that for each $y\in Y_1$, the LR counterpart $x \in X_1$ and $\Hat{x}\in\Hat{X}_1$ should not diverge too far.
We will refer to set $(\Hat{X}_1, Y_1)$ as the synthetic dataset.
With extra data pairs, we turns Eqn. \ref{mle} into
\begin{equation}
\label{map}
    \Hat{\theta} = \mathop{\arg\min}_{\theta} \sum_{(x,y)\in(\Hat{X}+\Hat{X}_1, \Hat{Y}+Y_1)}\mathcal{L}(f_{\theta}(x),y).
\end{equation}
During training the SR network $f_{\theta}$, we treat the synthetic data as additional training data and mix them with the original real data.
Both networks $f_{\theta}$ and $g_{\theta}$ have the same architecture (see Sec. \ref{arch}).
The main difference is that $g_{\theta}$ takes the HR image as input and generate its LR counterpart, while $f_{\theta}$ is modeling an inverse mapping.
The overall pipeline is shown in \figurename~\ref{fig:addis}.

This approach is mainly inspired by \textit{Back-Translation} \cite{sennrich2015improving,poncelas2018investigating} in Neural Machine Translation.
In the context of super resolution, \cite{bulat2018learn} proposes to use a GAN to stimulate image degradation and shares a similar motivation.
The fundamental differences between this paper and \cite{bulat2018learn} are two-fold: 1) we do not add any generative adversarial component into our PSNR-oriented models; 2) we train both networks with paired image data.

\subsection{Network Architecture}
\label{arch}
\begin{figure*}[t]
\begin{center}
   \includegraphics[width=1.0\linewidth]{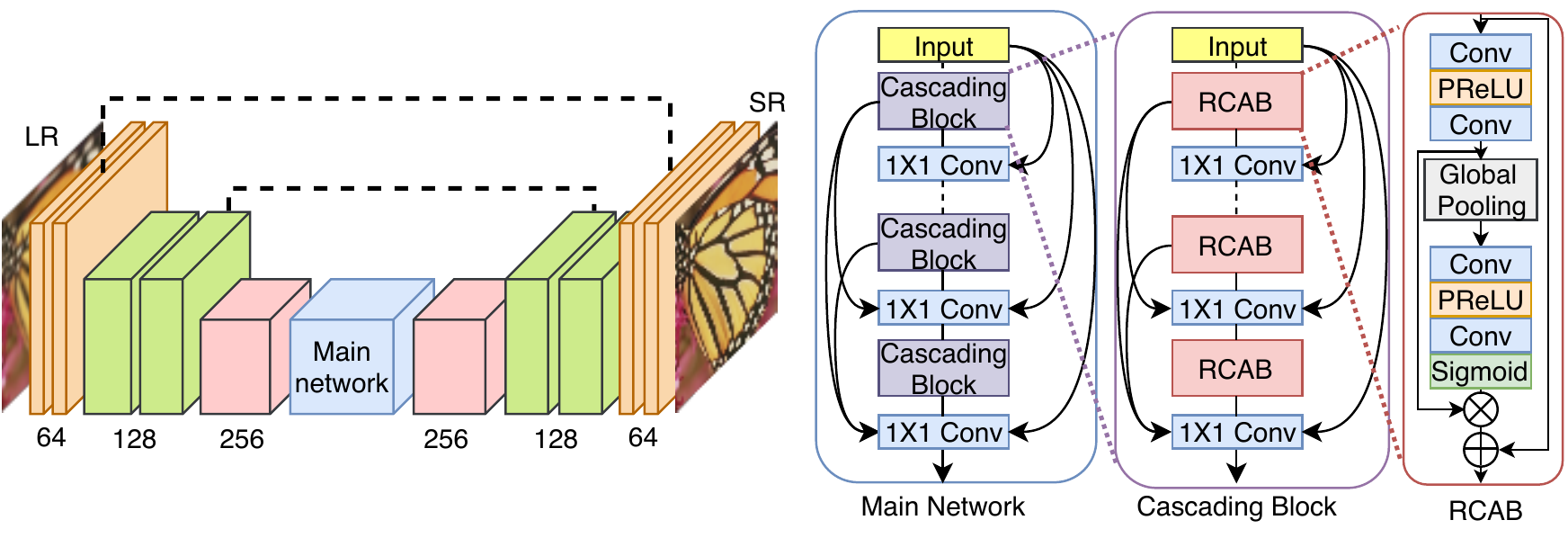}
\end{center}
\vskip -0.8cm
	\caption{Overall structure of our network.}
\label{fig:arch}
\vskip -0.4cm
\end{figure*}
As illustrated in \figurename~\ref{fig:arch}, the proposed network has a U-Net structure and consists of $4$ cascading blocks, each of which has $4$ Residual Channel Attention Blocks (RCABs).
The spatial resolution of features is decreased $2$ times using convolution layers with stride $2$, and then it is increased twice via pixel shuffle layers.
The basic building block is RCAB proposed in RCAN \cite{zhang2018image}, and the main difference between our model and RCAN is the global network topology.
Specifically, motivated by CARN \cite{ahn2018fast}, we use both local and global cascading modules to fully utilize hierarchical feature information derived from multiple blocks.
The outputs of RCAB are cascaded into higher layers, followed by a single $1\times1$ convolution layer, all of which serve as cascading blocks.
Similarly, global cascading modules have the same topology, where the unit blocks are replaced by cascading blocks.
To reduce computational cost, the main branch network works at $1/4H \times 1/4W$ resolution.

\section{Experiments}
\subsection{Technical Details}
For all experiments, we implement our models with the PyTorch \cite{2017-Paszke-p-} framework and train them using NVIDIA Titan Xp GPUs.
The mini-batch size is set to 16 and the spatial size of cropped patch is $128 \times 128$.
For initialization, the weights are randomly drawn from zero-mean Gaussian distributions as described in \cite{he2015delving}.
For optimization, we use Adam \cite{kingma2014adam} with $\beta_1 = 0.9$, $\beta_2 = 0.999$ and $\delta = 10^{-8}$.
The learning rate is initialized as $2\times10^{-4}$ and then decayed by half every $10^{5}$ iterations. We train all models for a total of $5\times10^5$ iterations. 
We use $\ell_1$ loss instead of $\ell_2$ as suggested in \cite{lim2017enhanced}.
We empirically set $\alpha=1.2$ for MixUp.
The SR results are evaluated on PSNR and SSIM \cite{wang2004image} on RGB space. For all convergence curves plotted in this paper, we calculate the average PSNR value on the central $1000\times1000$ patch of each image in validation set.

\subsection{Dataset}
\label{dataset}
We mainly train our models on the new Real-SR dataset, denoted as RealSR dataset below. The default splits of RealSR dataset consist of $60$ training images, $20$ validation images and $20$ test images. Evaluation of the trained models is performed on $20$ validation images since test images are not publicly available. As described in Sec. \ref{syn}, we also include a prevalent DIV2K dataset \cite{Agustsson_2017_CVPR_Workshops} as additional training data, since these images cover diverse contents, including objects, environments, animals, natural scenery, etc. Following \cite{lim2017enhanced}, we use $800$ training images as training set.

To prepare training data, we first crop the HR images into a set of $480\times480$ sub-images with a stride $240$ for DIV2K dataset.
Similarly, we crop HR images into sub-images of size $200\times200$ and stride $100$ for RealSR dataset.
In this manner we have totally $12,837$ and $32,208$ sub-images from RealSR and DIV2K dataset, respectively.
To fully utilize the dataset, training images are augmented with random horizontal/vertical flips and rotations.
During training, a patch of size $128\times128$ is randomly cropped from a sub-image.

\subsection{Experiments on MixUp}
\label{exp:mixup}
\begin{figure}[t]
\begin{center}
   \includegraphics[width=1.0\linewidth]{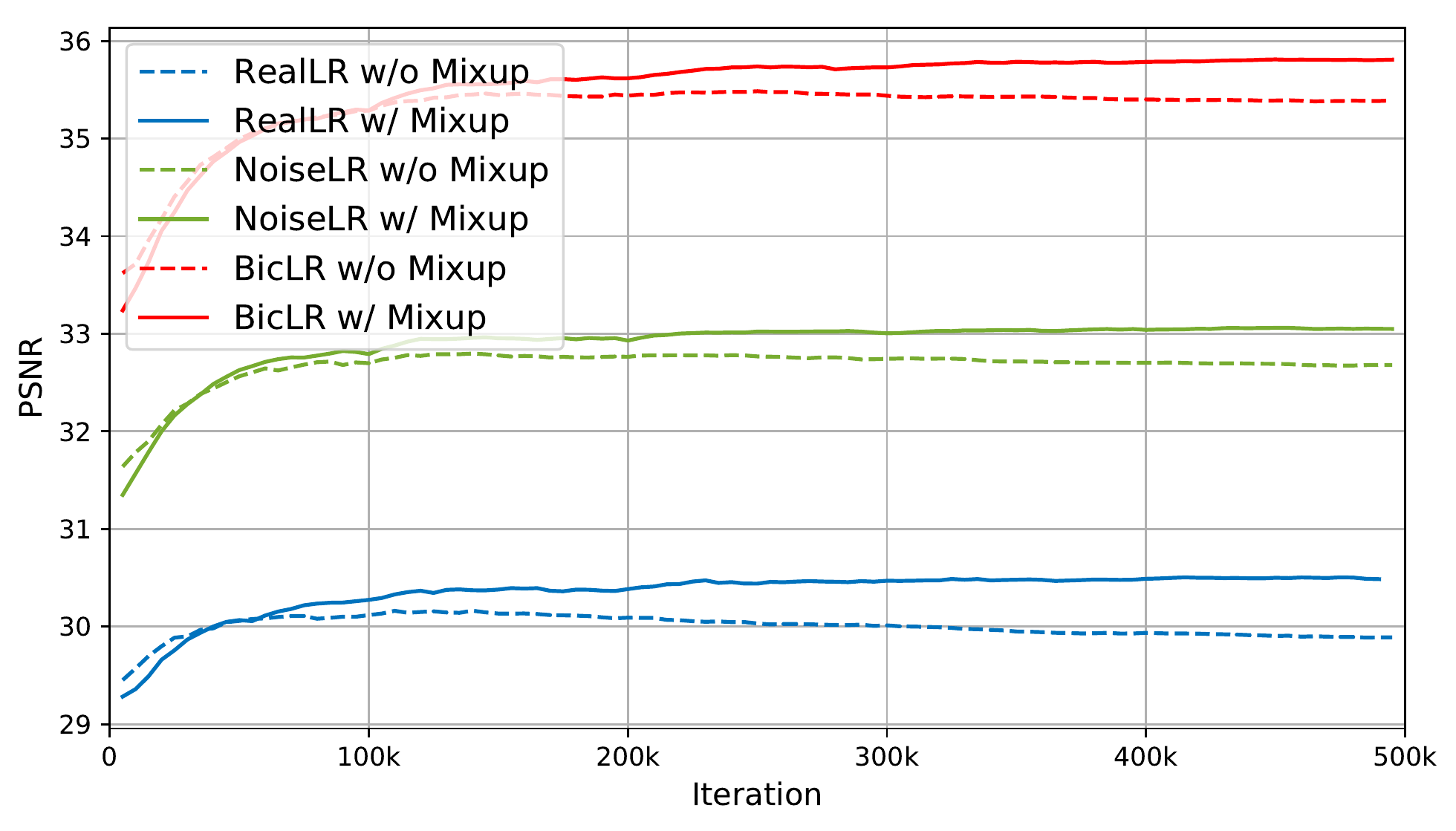}
\end{center}
\vskip -0.6cm
   \caption{Convergence curves of models trained w/ and w/o MixUp. ``RealLR'', ``NoisyLR'' and ``BicLR'' indicate images sampled from $\Hat{X}_{real}$, $\Hat{X}_{noise}$ and $\Hat{X}_{bic}$.}
\label{fig:mixup}
\vskip -0.2cm
\end{figure}
In this section we study the effect of MixUp on different types of dataset. Different from Sec. \ref{exp:aug}, we only use $12,837$ sub-images from RealSR dataset as training set.
As described in Sec. \ref{mixup}, MixUp serves as a regularization on data manifold.
To verify the effectiveness of this regularization on various types of degradation, we study three settings by generating LR from HR images as follows:
\begin{itemize}
\item Real LR images from RealSR training set
\item Bicubic downsample HR images with a factor $4\times$ and then upsample to the original resolution.
\item Bicubic downsample HR images with a factor $4\times$ and then upsample to the original resolution, with realistic noise \cite{guo2018toward} added to LR images. 
\end{itemize}
Similarly, the corresponding validation set is constructed in the same manner for each setting.
We denote the LR images as $\Hat{X}_{real}$, $\Hat{X}_{bic}$ and $\Hat{X}_{noise}$, which have the same ground truth $\Hat{Y}$.
On three datasets we train models with and without MixUp to investigate effects of MixUp.

It can be observed from \figurename~\ref{fig:mixup} that after the first learning rate decay ($100$K), models trained without MixUp quickly deteriorate their validation performance due to overfitting, while those with MixUp keep the same validation accuracy until termination.
In super-resolution task, MixUp significantly reduces overfitting and guarantees robust training.

\subsection{Experiments on Data Synthesis}
\label{exp:aug}
\begin{figure}[t]
\begin{center}
   \includegraphics[width=1.0\linewidth]{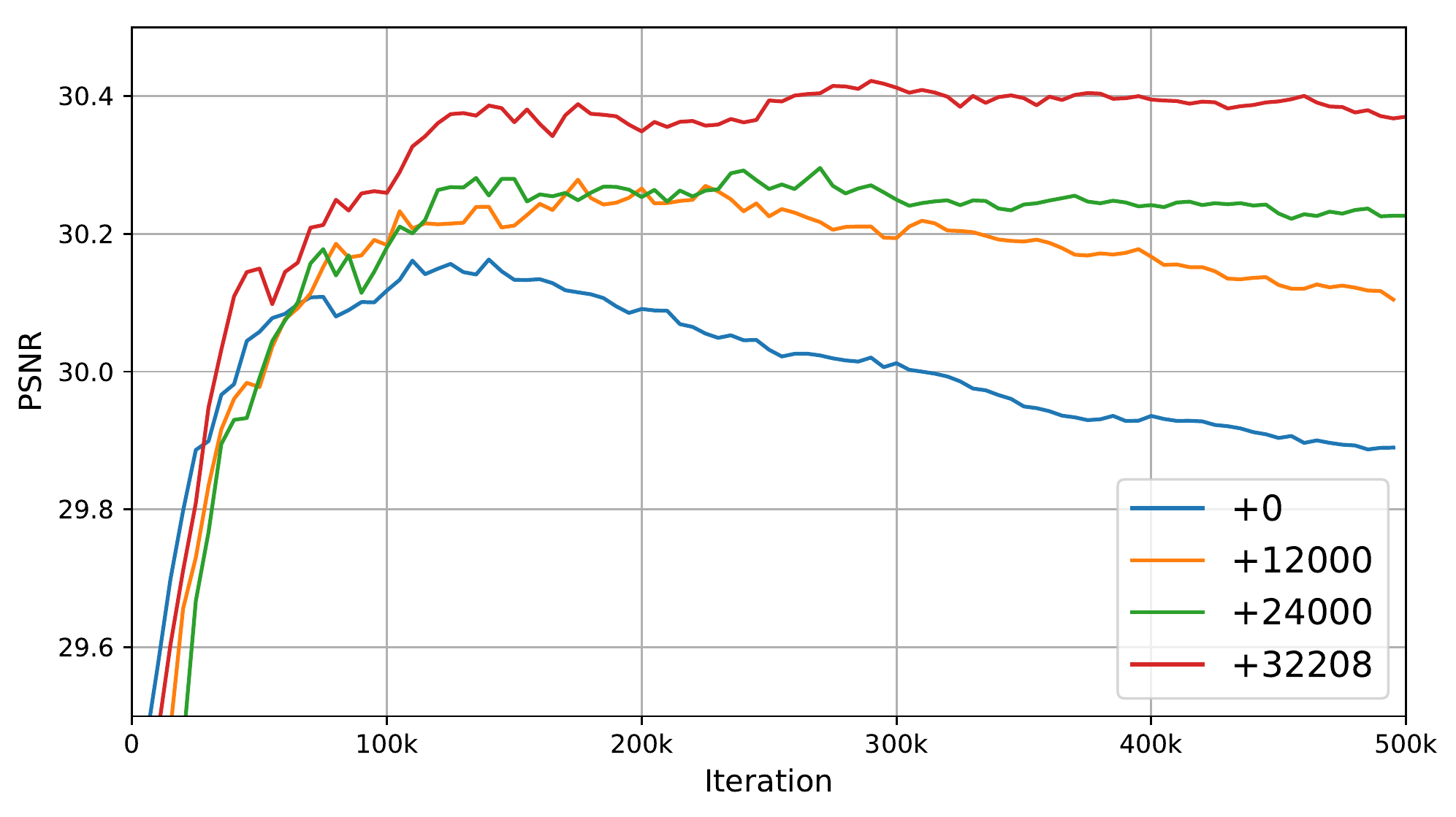}
\end{center}
\vskip -0.6cm
   \caption{Convergence curves of SR networks trained on observed data combined with different amounts of synthetic data.}
\label{fig:diversity}
\vskip -0.2cm
\end{figure}

In the scope of this section, we mainly use $12,837$ sub-image pairs from RealSR dataset as the observation set and $32,208$ HR sub-images from DIV2K dataset for data synthesis.
We first train the degradation model $g_{\theta}$ with $12,837$ training sub-image pairs and the training settings are same as those for $f_{\theta}$.
The model converges at $20$K iterations.
We aim to provide a systematic analysis of SR networks trained on different synthetic dataset $(\Hat{X}_1, Y_1)$ to build a clearer picture about the progressive effects of incremental amounts of synthetic data to the generalization ability.

To validate the assumption that the observation set $(\Hat{X}, \Hat{Y})$ is biased sampled, we evaluate how the validation error varies while increasing volumes of synthetic data (i.e., higher diversity).
Specifically, models are built using a base observation set combined with the augmentation set $(\Hat{X}_1, Y_1)$ that starts with $0$ sub-image and grows incrementally to all $32,208$ sub-images.
Note that the experimental settings degenerate to a baseline scenario without any regularization when $(\Hat{X}_1, Y_1)$ contains no sub-image.

According to the results shown in \figurename~\ref{fig:diversity}, the benefits of adding synthetic data are delaying and reducing overfitting on training set.
As expected, adding more and more synthetic data to the training set encourages better generalization.
The best combination comprises $45,045$ sub-images ($12,837$ from $(\Hat{X}, \Hat{Y})$ and $32,208$ from $(\Hat{X}_1, Y_1)$), which achieves a PSNR of $30.46$dB, $0.25$dB better than the baseline model.

\subsection{Comparison with the State-of-the-arts}
To further investigate overfitting on limited data, we include both light-weight networks (e.g., FSRCNN \cite{dong2016accelerating}, CARN \cite{ahn2018fast}) and larger networks (e.g., RCAN \cite{zhang2018image}, RRDB \cite{wang2018esrgan}) in our comparison.
We reimplement these state-of-the-art methods on RealSR dataset.
Note that most of the existing methods operate at low resolution and upsample feature maps at the very end of the networks.
Therefore, we simply modify the models by downsampling LR images with a stride $4$ in the first convolution layer, which is consistent with our U-Net architecture.
Throughout experiments, we find existing large models can easily overfit to the training set, and therefore we study early stopped versions of those models to provide a stronger comparison.
In contrast, early stopping is not necessary for light-weight networks and our method.
We stress that early stopping strategy does not solve the overfitting problem (see also Sec. \ref{overfit}), as both training error and validation error are high.
With early stopping, a large model will underfit and fail to make full use of model capacity.
Specifically, an early stopped large model tends to restore blurry images while a overfitted version generates sharp images with unpleasant artifacts.
Following \cite{lim2017enhanced}, self-ensemble strategy is also applied to further improve generalization performance and the self-ensemble version is denoted with ``*''.

\tablename~\ref{tab:sota} lists the quantitative results (PSNR / SSIM) on RealSR validation set.
These results provide two insights: (1) both MixUp and data synthesis can significantly suppress overfitting on limited training data. (2) MixUp and data synthesis are not mutually exclusive, as one can additionally apply MixUp technique on the additional synthetic data to further improve the final performance.

\begin{figure*}[t]
\centering
    \begin{subfigure}[t]{1.0\linewidth}
        \centering
        \includegraphics[width=0.85\linewidth]{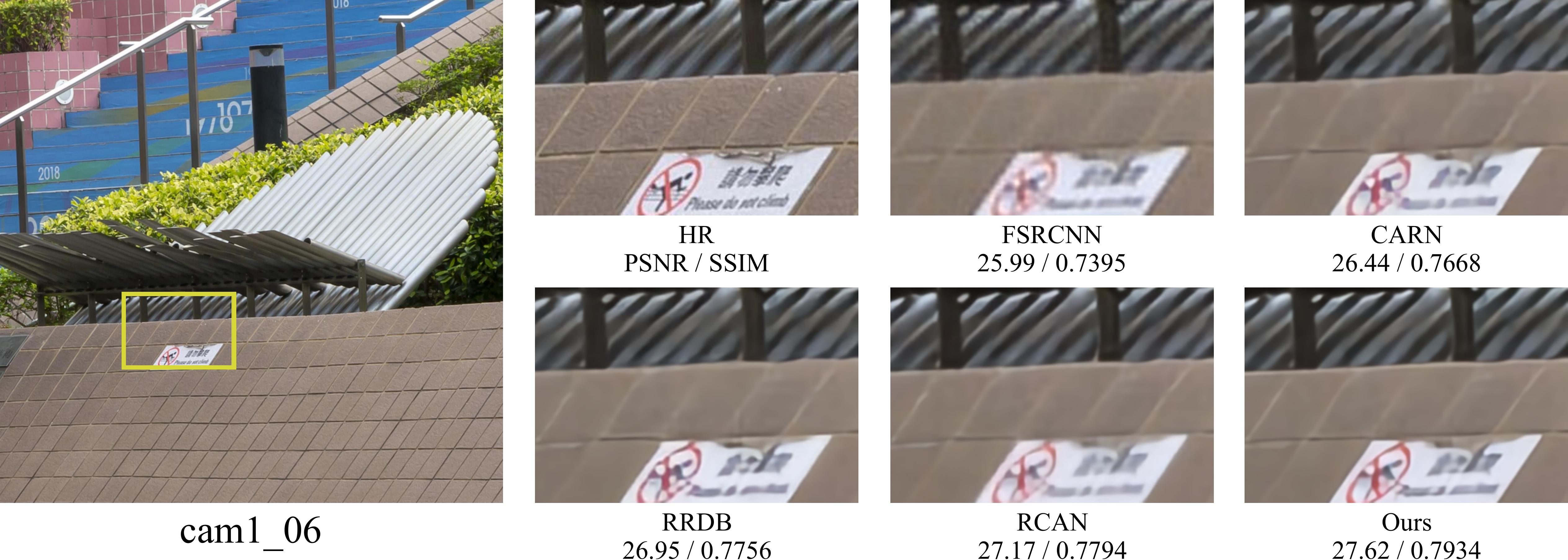}
    \end{subfigure}\\
    \begin{subfigure}[t]{1.0\linewidth}
        \centering
        \includegraphics[width=0.85\linewidth]{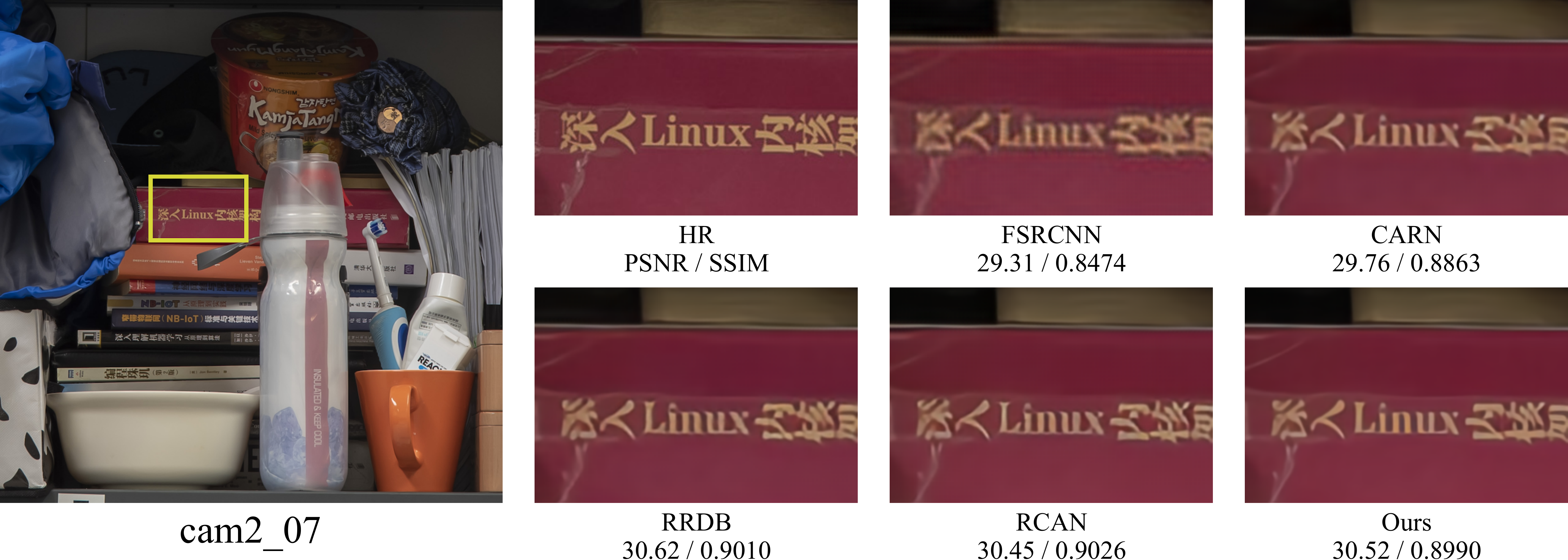}
    \end{subfigure}
    \begin{subfigure}[t]{1.0\linewidth}
        \centering
        \includegraphics[width=0.85\linewidth]{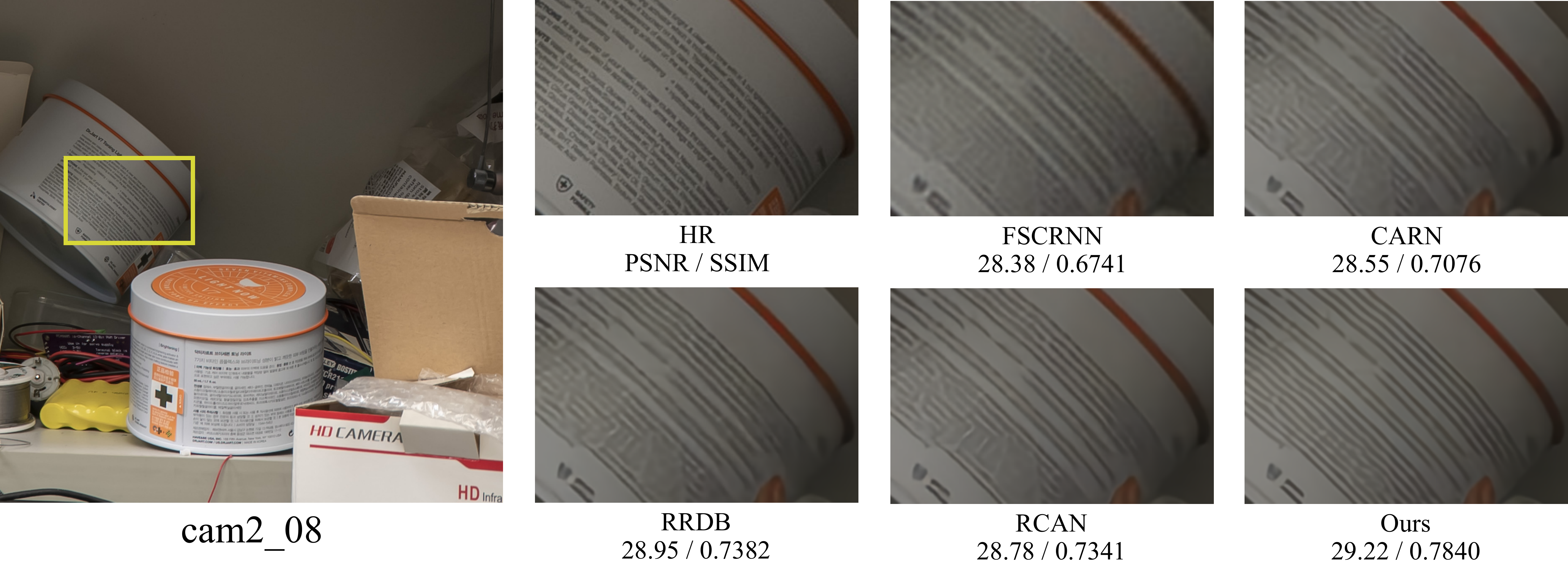}
    \end{subfigure}
\caption{Visual comparison of FSRCNN \cite{dong2016accelerating}, CARN \cite{ahn2018fast}, RRDB \cite{wang2018esrgan}, RCAN \cite{zhang2018image} and our method on validation dataset.}
\label{fig:vusial}
\vskip -0.4cm
\end{figure*}

In \figurename~\ref{fig:vusial}, we show visual comparisons on state-of-the-art networks and our model. For image ``cam2\_08'', we observe that most of the compared methods cannot recover the lines of text and would suffer from blurring artifacts. In contrast, our model can alleviate the blurring artifacts better and recover more details. Similar observations are shown in images ``cam2\_07'' and ``cam1\_06''.

\begin{table}
\small
\begin{center}
\caption{Model comparisons on  validation set. The best and second best results are \textbf{highlighted} and \underline{underlined}, respectively. ``+ES'' denotes early stopping and ``*'' denotes self-ensemble strategy.}
\label{tab:sota}
\begin{tabular}{ccc}
\hline
Method & PSNR & SSIM\\
\hline
FSRCNN\cite{dong2016accelerating} &$28.3394$ &$0.8254$ \\
CARN\cite{ahn2018fast} + ES &$29.1620$ &$0.8580$ \\
RRDB\cite{wang2018esrgan} + ES &$29.4581$ &$0.8643$ \\
RCAN\cite{zhang2018image} + ES &$29.6299$ &$0.8675$ \\\hline
U-Net(Ours) + Synthesis &$29.8503$ &$0.8731$ \\
U-Net(Ours) + MixUp &$29.9055$ &$0.8729$ \\
U-Net(Ours) + Synthesis + MixUp &$\underline{30.0278}$ &$\underline{0.8753}$ \\
U-Net(Ours)* + Synthesis + MixUp &$\textbf{30.1624}$ &$\textbf{0.8777}$ \\
\hline
\end{tabular}
\end{center}
\vskip -0.4cm
\end{table}

\section{Discussion}
\begin{figure}[t]
\begin{center}
   \includegraphics[width=1.0\linewidth]{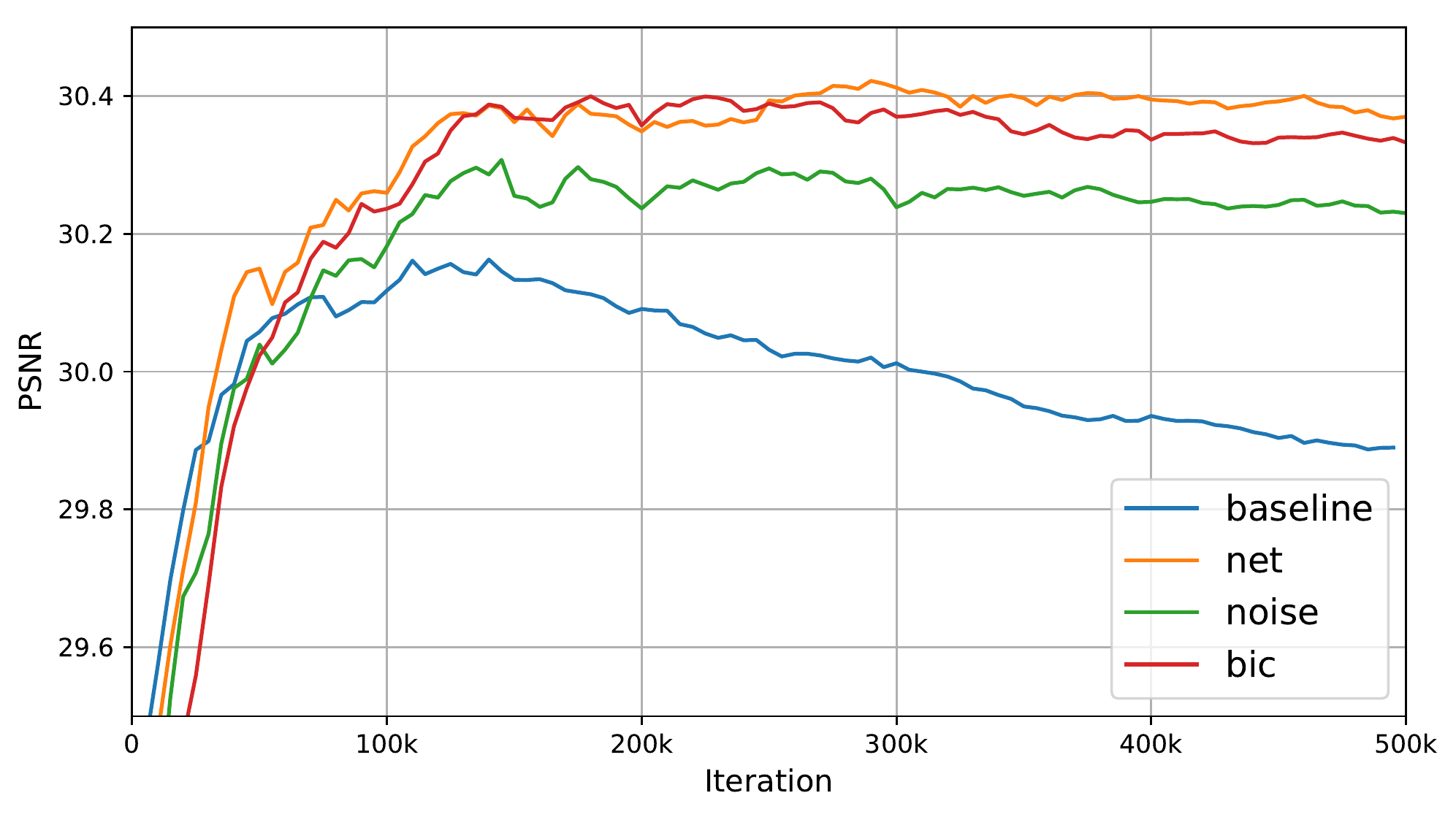}
\end{center}
\vskip -0.6cm
   \caption{Convergence curves of SR networks trained on observed data combined with different types of synthetic data.}
\label{fig:similarity}
\vskip -0.4cm
\end{figure}

In this section we further discuss the effectiveness of data synthesis.
With a sufficiently large dataset comprising high-quality HR images, one question remains unanswered is how the quality of generated LR images affects generalization ability.
Our investigation involves applying various degradation types to HR images from DIV2K training set, while RealSR dataset remains unchanged.
LR images are produced with three different degradation processes:
\begin{itemize}
\item Add White Gaussian noise with $\sigma=25$ to HR images.
\item Bicubic downsample HR images with a factor $4\times$ and then upsample to the original resolution.
\item Construct a network to learn degradation.
\end{itemize}
The corresponding data pairs constitute a synthetic dataset, where we will refer to these augmentation set as $\Hat{X}_{1\_noise}$, $\Hat{X}_{1\_bic}$ and $\Hat{X}_{1\_net}$.
Convergence curves of models trained on different types of augmentation set are shown in \figurename~\ref{fig:similarity}.
We see that the use of synthetic data essentially reduce overfitting problem, compared with the baseline.
In addition, LR images from $\Hat{X}_{1\_noise}$, $\Hat{X}_{1\_bic}$ and $\Hat{X}_{1\_net}$ are completely different from each other.
The best generalization is reached by the model trained with $\Hat{X}_{1\_net}$, indicating that the learned mapping function $g_{\theta}$ among the investigated degradation types would be the most ``similar'' to the unknown true degradation $g$.
One can also investigate the sensitivity of SR networks to different kinds of degradation models, which will be left to our future work.

\section{Conclusion}
In this paper, we propose two simple yet effective methods to reduce overfitting problem in SR networks.
Our method won the second place in NTIRE2019 Real SR Challenge.
Particularly, we introduce MixUp technique to encourage networks trained with limited data to generalize well.
In addition, data synthesis with learned degradation are employed to train models using extra high-quality images with higher content diversity.
This strategy proves to be successful in reducing biases of data.
By combining both techniques, large models can be trained without overfitting and achieve satisfactory generalization performance.
Since the proposed approach is network-independent, it is expected to be easily applied to other network architectures and image restoration tasks. Future work will explore the effectiveness of our approach in more settings.

\textbf{Acknowledgements.} This work is partially supported by National Key Research and Development Program of China (2016YFC1400704), Shenzhen Research Program (JCYJ20170818164704758, JCYJ20150925163005055, CXB201104220032A), and Joint Lab of CAS-HK.
{\small
\bibliographystyle{ieee_fullname}
\bibliography{egbib}
}
\end{document}